# A Bayesian Method for Constructing Bayesian Belief Networks from Databases

Gregory F. Cooper
Section of Medical Informatics
Department of Medicine
University of Pittsburgh
Pittsburgh, PA 15261

Edward Herskovits
Medical Computer Science
Stanford University
Stanford, CA 94305

## Abstract

This paper presents a Bayesian method for constructing Bayesian belief networks from a database of cases. Potential applications include computer-assisted hypothesis testing, automated scientific discovery, and automated construction of probabilistic expert systems. Results are presented of a preliminary evaluation of an algorithm for constructing a belief network from a database of cases. We relate the methods in this paper to previous work, and we discuss open problems.

# 1 INTRODUCTION

Decision making typically is replete with uncertainty. In general, it is important that computer systems that assist in decision making be capable of representing and reasoning with uncertainty. Probabilistic networks provide a precise and concise representation of probabilistic dependencies among variables. In the last few years, significant progress has been made in formalizing the theory of probabilistic networks [Pearl 1988]. Advances also have occurred in improving the efficiency of methods for inference on probabilistic networks [Henrion 1990], although for some complex networks additional improvements are still needed. The feasibility of using probabilistic networks in constructing diagnostic systems has been demonstrated in several domains [Agogino and Rege 1987, Andreassen, Woldbye, et al. 1987, Beinlich, Suermondt, et al. 1989, Heckerman, Horvitz, et al. 1989, Henrion and Cooley 1987, Holtzman 1989, Suermondt and Amylon 1989].

Although substantial advances have been made in developing the theory and application of probabilistic networks, the actual construction of these networks often remains a difficult, time-consuming task. The task is time-consuming because typically it must be performed manually by an expert or with an expert. Some important progress has been made in developing methods to improve the efficiency of knowledge acquisition from experts [Heckerman 1990]. These methods are likely to remain important in domains of small to moderate size in which there are readily available experts. Some domains, however, are large. In others, there are few, if any, available experts. Methods for assisting, or in some cases replacing, the manual expert-based methods of knowledge acquisition are needed.

Databases are becoming increasingly abundant in many areas, including science, engineering, and business. In each of these areas, there are many potential opportunities for using probabilistic networks to provide assistance in decision making. By using databases to assist in constructing probabilistic networks, we may be able to decrease knowledge acquisition time significantly. Automatically generated networks could be used directly to provide decision-making assistance, or used as a starting point for modification by an expert. In the latter case, the editing of a network may require substantially less time than de novo generation of the network by an expert.

The automated construction of probabilistic networks also can provide insight into the probabilistic dependencies that exist among the domain variables. One application is the automated discovery of dependency relationships. The computer program searches for a probabilistic network structure that has a high posterior probability given the database, and outputs the structure and its probability. A related task is computer-assisted hypothesis testing: the user enters a hypothesized structure of the dependency relationships among a set of variables and the program calculates the probability of the structure given a database of cases on the variables. These applications have the potential to affect broad areas of scientific discovery and data evaluation.

As an example, consider the fictitious database of cases shown in Table 1. Suppose that $x_1$ is an experimental condition and $x_2$ and $x_3$ are two experimental outcomes.

Given the database, what are the qualitative dependency relationships among the variables? For example, do $x_1$ and $x_3$ influence each other directly, or do they do so only through $x_2$? What is the probability that $x_3$ will be present if $x_1$ is present? Clearly, there are no categorically correct answers to each of these questions. The answers depend on a number of factors, including the model that we use to represent the data, and our prior knowledge about the data in the database and the relationships among the variables. In this paper, we do not attempt to consider all such factors in their full generality. Rather, we specialize the general task by presenting one particular framework for constructing probabilistic networks from databases, as for example the database in Table 1, such that these networks can be used for probabilistic inference, as for instance in calculating $P(x_3 = \text{present} \mid x_1 = \text{present})$. In particular, we focus on using a Bayesian belief network as a model of probabilistic dependency. Our primary goal is to construct such a network (or networks), given a database and a set of explicit assumptions about our prior probabilistic knowledge of the domain, and then use that network (or networks) for inference.

Table 1: A database example. For notational convenience, in the text we sometimes use 0 to denote *absent* and 1 to denote *present*.

| Case | Variable values for each case | | |
|------|---------|---------|---------|
|      | $x_1$ | $x_2$ | $x_3$ |
| 1 | present | absent | absent |
| 2 | present | present | present |
| 3 | absent | absent | present |
| 4 | present | present | present |
| 5 | absent | absent | absent |
| 6 | absent | present | present |
| 7 | present | present | present |
| 8 | absent | absent | absent |
| 9 | present | present | present |
| 10 | absent | absent | absent |

# 2 METHODS

In this section, we first briefly review some key concepts about Bayesian belief networks. Then, we present the primary theoretical developments of our work thus far in developing methods for learning the structure of Bayesian belief networks from databases; a more detailed discussion with proofs appears in [Cooper and Herskovits 1991]. Finally, we discuss the empirical results of an algorithm that applies this theory to search for the most likely belief-network structure, given a database.

## 2.1 BAYESIAN BELIEF NETWORKS

A Bayesian belief-network *structure* $B_S$ is a directed acyclic graph in which nodes represent domain variables and arcs between nodes represent probabilistic dependencies [Cooper 1989, Horvitz, Breese, et al. 1988, Lauritzen and Spiegelhalter 1988, Neapolitan 1990, Pearl 1986, Pearl 1988, Shachter 1988]. A variable in a Bayesian belief-network structure may be continuous [Shachter and Kenley 1989] or discrete. In this paper, we shall focus our discussion on discrete variables. Figure 1a shows an example of a belief-network structure, which we shall call $B_{S1}$, containing three variables. The arc from $x_1$ to $x_2$ indicates that these two variables are probabilistically dependent. Similarly, the arc from $x_2$ to $x_3$ indicates a probabilistic dependency between these two variables. The absence of an arc from $x_1$ to $x_3$ implies that there is no direct probabilistic dependency between $x_1$ and $x_3$. In particular, the probability of each value of $x_3$ is conditionally independent of the value of $x_1$ *given* that the value of $x_2$ is known. Figure 1b shows an alternative structure that expresses different dependency relationships among the three variables. The representation of conditional dependencies and independencies is the essential function of belief networks. For a detailed discussion of the semantics of Bayesian belief networks, see [Pearl 1988].

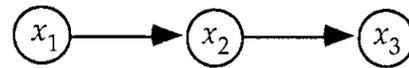

(a) Structure $B_{S1}$

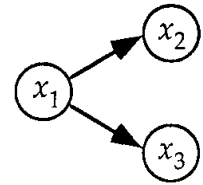

(b) Structure $B_{S2}$

Figure 1: Two alternative belief-network structures on three variables.

A Bayesian belief-network structure $B_S$ is augmented by conditional probabilities, $B_P$, to form a Bayesian belief network $B$. Thus, $B = (B_S, B_P)$. For brevity, we call $B$ a belief network. For each node[1] in the network structure, there is a conditional probability function that relates this node to its immediate predecessors (parents). We shall use $\pi_i$ to denote the parent nodes of variable $x_i$. If a node has no parents, then a prior probability function, $P(x_i)$, is specified. A set of probabilities is shown in Table 2 for the belief-network structure in Figure 1a. We shall use the term *conditional probability* to refer to a probability statement, such as $P(x_2 = \text{present} \mid x_1 = \text{present})$. We use the term *conditional probability assignment* to denote a numerical assignment to a conditional probability, as, for example, the assignment $P(x_2 = \text{present} \mid x_1 = \text{present}) = 0.8$. The network structure $B_{S_1}$ in Figure 1a and the probabilities $B_{P_1}$ in Table 2 together define a belief network, which we denote as $B_1$.

Table 2: The probability assignments associated with the belief-network structure $B_{S_1}$ in Figure 1. We shall denote these probability assignments as $B_{P_1}$.

$P(x_1 = \text{present}) = 0.6$
$P(x_1 = \text{absent}) = 0.4$
$P(x_2 = \text{present} \mid x_1 = \text{present}) = 0.8$
$P(x_2 = \text{absent} \mid x_1 = \text{present}) = 0.2$
$P(x_2 = \text{present} \mid x_1 = \text{absent}) = 0.3$
$P(x_2 = \text{absent} \mid x_1 = \text{absent}) = 0.7$
$P(x_3 = \text{present} \mid x_2 = \text{present}) = 0.9$
$P(x_3 = \text{absent} \mid x_2 = \text{present}) = 0.1$
$P(x_3 = \text{present} \mid x_2 = \text{absent}) = 0.15$
$P(x_3 = \text{absent} \mid x_2 = \text{absent}) = 0.85$

Belief networks can be used to represent the probabilities over any discrete sample space: the probability of any sample point in that space can be computed from the probabilities in the belief network. The key feature of belief networks is their explicit representation of the conditional independence among events. In particular, investigators have shown [Kiiveri, Speed, et al. 1984, Pearl 1988, Shachter 1986] that the joint probability of any particular instantiation[2] of all $n$ variables in a belief network can be calculated as

$$P(X_1, \ldots, X_n) = \prod_{i=1}^{n} P(X_i \mid \pi_i), \qquad (1)$$

[1] Since there is a one-to-one correspondence between a node in $B_S$ and a variable in $B_P$, we shall use the terms *node* and *variable* interchangeably.

[2] An *instantiated variable* is a variable with an assigned value. When we need to designate a particular value $v_{ik}$ of variable $x_i$, we shall write $x_i = v_{ik}$.

where each $X_i$ represents an instantiated variable and $\pi_i$ represents an instantiation of the parents of $X_i$.

Therefore, the joint probability of any instantiation of all the variables in a belief network can be computed as the product of only $n$ probabilities. In principle, we can recover the complete joint-probability space from the belief-network representation by calculating the joint probabilities that result from every possible instantiation of the $n$ variables in the network. Thus, we can determine any probability of the form $P(Z \mid Y)$, where $Z$ and $Y$ are sets of variables with known values (instantiated variables). For example, for our sample three-node belief network $B_1$, $P(x_3 = \text{present} \mid x_1 = \text{present}) = 0.75$.

Let us now consider the problem of finding the most probable belief-network structure, given a database. Once such a structure is found, we can derive numerical belief-network probabilities from the database [Cooper and Herskovits 1991]. We can use the resulting belief network for probabilistic inference, such as calculating the value of $P(x_3 = \text{present} \mid x_1 = \text{present})$. In addition, the structure may lend insight into the dependency relationships among the variables in the database, as, for example, possible causal relationships.

To be more specific, let $D$ be a database of cases, $Z$ be the set of variables represented by $D$, and $B_{S_i}$ and $B_{S_j}$ be two belief-network structures containing exactly those variables that are in $Z$. In the next section, we develop a method for computing $P(B_{S_i} \mid D)/P(B_{S_j} \mid D)$. By computing such ratios for pairs of belief-network structures, we can rank order a set of structures by their posterior probabilities. To calculate the ratio of posterior probabilities, we shall calculate $P(B_{S_i}, D)$ and $P(B_{S_j}, D)$ and use the following equivalence:

$$\frac{P(B_{S_i} \mid D)}{P(B_{S_j} \mid D)} = \frac{\dfrac{P(B_{S_i}, D)}{P(D)}}{\dfrac{P(B_{S_j}, D)}{P(D)}} = \frac{P(B_{S_i}, D)}{P(B_{S_j}, D)}. \qquad (2)$$

## 2.2 A FORMULA FOR COMPUTING $P(B_S, D)$

Let $B_S$ represent an arbitrary belief-network structure containing just the variables in $Z$. In this section, we present a method for calculating $P(B_S, D)$. In doing so, we shall introduce five assumptions that render this calculation computationally tractable.

Assumption 1. The process that generated the database is modeled as a belief network containing just the variables in $Z$, which are discrete.

As this assumption states, we shall not consider continuous variables in this paper.

A belief network (structure *plus* conditional probabilities) is sufficient to capture any probability distribution over the variables in $Z$ [Pearl 1988]. A belief-network structure alone, containing just the variables in $Z$, can capture many—but not all—the qualitative independence relationships that might exist in an arbitrary probability distribution over $Z$ [Pearl 1988]. Assumption 1, therefore, is justified to the extent that the relationships of independence and dependence, among variables in the underlying process that is hidden from us, can be represented by some belief network. In the remainder of this section, we shall write as though database $D$ was generated by Monte Carlo sampling of a belief network with structure $B_S$ that is hidden from us. One of our primary goals will be to use $D$ to try to discover $B_S$. In this section, we assume that $B_S$ contains just the variables in $Z$. In [Cooper and Herskovits 1991], we allow $B_S$ to contain variables beyond those in $Z$.

The application of Assumption 1 yields

$$P(B_S, D) = \int_{B_P} P(D \mid B_S, B_P) f(B_P \mid B_S) P(B_S) \, dB_P, \qquad (3)$$

where $B_P$ is a vector whose values denote the conditional-probability assignments associated with belief-network structure $B_S$, and $f$ is the conditional probability density function over $B_P$ given $B_S$. The integral is over all possible value assignments to $B_P$. Thus, we are integrating over all possible belief networks that can have structure $B_S$. The integral in Equation 3 represents a multiple integral and the variables of integration are the conditional probabilities associated with structure $B_S$.

Assumption 2. Cases occur independently, given a belief network model.

This assumption is equivalent to assuming that the belief network that is generating the data is static, that is, it does not change as cases are being generated. It follows from the conditional independence of cases expressed in Assumption 2 that Equation 3 may be rewritten as

$$P(B_S, D) = P(B_S) \int_{B_P} \left[ \prod_{j=1}^{m} P(C_j \mid B_S, B_P) \right] f(B_P \mid B_S) \, dB_P, \qquad (4)$$

where $m$ is the number of cases in $D$ and $C_j$ is the $j$th case in $D$.

Assumption 3. Cases are complete, that is, there are no cases that have variables with missing values.

In [Cooper and Herskovits 1991], we relax this assumption. We now introduce additional notation to facilitate the application of Assumption 3. Let $D_{ij}$ denote the value assignment of variable $i$ in case $j$. Thus, $D_{21} = 0$, since $x_2 = 0$ (i.e., $x_2$ is absent) in case 1 in Table 1. In $B_S$, for every variable $x_i$, there is a set of parents $\pi_i$ (possibly the empty set). For each case in $D$, the variables in $\pi_i$ are each assigned a particular value; call such a case-specific instantiation of all the variables in $\pi_i$ a $\pi$-instantiation. Let $\phi_i$ denote a list of the unique $\pi$-instantiations for the parents of $x_i$ as seen in $D$. If $x_i$ has no parents, then we define $\phi_i$ to be the list $(\varnothing)$, where $\varnothing$ represents the empty set of parents. Although the ordering of the elements of $\phi_i$ is arbitrary, we shall use a list (vector), rather than a set, so that we can refer to members of $\phi_i$ using an index. For example, consider variable $x_2$ in $B_{S1}$, which has parent set $\pi_2 = \{x_1\}$. In this example, $\phi_2 = ((x_1 = 0), (x_1 = 1))$, because there are cases in $D$ where $x_1$ has the value 0 and cases where it has the value 1. Let $\phi_i[j]$ be the $j$th element of $\phi_i$. Thus, for example, $\phi_2[1]$ is equal to $(x_1 = 0)$. Let $\sigma(i, j)$ be an index function, such that the instantiation of $\pi_i$ in case $j$ is the $\sigma(i, j)^{\text{th}}$ element of $\phi_i$. Thus, for example, $\sigma(2, 1) = 2$, because in case 1 the parent set of variable $x_2$ — namely $\{x_1\}$ — is instantiated as $x_1 = 1$, which is the second element of $\phi_2$. Therefore, $\phi_2[\sigma(2, 1)]$ is equal to $(x_1 = 1)$. Let $q_i = |\phi_i|$. Note that since there are $m$ cases in $D$, $q_i \le m$. Since, according to Assumption 3, cases are complete, we can apply Equation 1 to represent the probability of a case as follows:

$$P(C_j \mid B_S, B_P) = \prod_{i=1}^{n} P(x_i = D_{ij} \mid \phi_i[\sigma(i,j)], B_P) \qquad (5)$$

Substituting Equation 5 into Equation 4, we obtain

$$P(B_S, D) = P(B_S) \int_{B_P} \left[ \prod_{j=1}^{m} \prod_{i=1}^{n} P(x_i = D_{ij} \mid \phi_i[\sigma(i,j)], B_P) \right] \times f(B_P \mid B_S) \, dB_P. \qquad (6)$$

For a given $i$ and $j$, let $f(P(x_i \mid \phi_i[j], B_P))$ denote our probability distribution over the possible values of $P(x_i \mid \phi_i[j], B_P)$. That is, the density function $f(P(x_i \mid \phi_i[j], B_P))$ represents our belief about the values to assign the conditional probability function $P(x_i \mid \phi_i[j], B_P)$. For notational convenience, we shall leave the term $B_P$ implicit. We shall assume that our belief about the values to assign to a conditional probability function in a belief network is not influenced by our belief about the values to assign any other conditional probability function. More formally, we can express this assumption as follows:

<u>Assumption 4.</u> For $1 \le i, i' \le n$, $1 \le j \le q_i$, $1 \le j' \le q_{i'}$, if $ij \ne i'j'$ then the distribution $f(P(x_i \mid \phi_i[j]))$ is marginally independent of the distribution $f(P(x_{i'} \mid \phi_{i'}[j']))$.

As an example, Assumption 4 implies that our belief about the assignment of a value to the conditional probability $P(x_3 = 0 \mid x_2 = 0)$ is independent of our assignment of a value to the conditional probability $P(x_2 = 0 \mid x_1 = 0)$, since these two probabilities are components of different conditional probability distributions. However, our belief about $P(x_2 = 0 \mid x_1 = 0)$ must be dependent on our belief about $P(x_2 = 1 \mid x_1 = 0)$, since they are members of the same conditional probability distribution; in particular, since $x_2$ is a binary variable, $P(x_2 = 0 \mid x_1 = 0) = 1 - P(x_2 = 1 \mid x_1 = 0)$.

<u>Assumption 5.</u> For $1 \le i \le n$, $1 \le j \le q_i$, the distribution $f(P(x_i \mid \phi_i[j]))$ is uniform.

This assumption states that initially, before we see the data, we are indifferent regarding giving one assignment of values to a conditional probability function versus some other assignment. This probability density function is, however, just a special case of the Dirichlet distribution [deGroot 1970]. In [Cooper and Herskovits 1991] we generalize Assumption 5 by representing $f(P(x_i \mid \phi_i[j]))$ with a Dirichlet distribution.

Assumptions 4 and 5 permit us to define the joint density function $f(B_P \mid B_S)$ using density functions of the form $f(P(x_i \mid \phi_i[j]))$ which are uniform.

We now use Assumptions 1 through 5 in the following theorem, proven in [Cooper and Herskovits 1991], which solves Equation 3 for $P(B_S, D)$:

**Theorem** Let $Z$ be a set of $n$ discrete variables, where a variable $x_i$ in $Z$ has $r_i$ possible value assignments: $(v_{i1}, \ldots, v_{ir_i})$. Let $D$ be a database of $m$ cases, where each case contains a value assignment for each variable in $Z$. Let $B_S$ denote a belief-network structure containing just the variables in $Z$. Each variable $x_i$ in $B_S$ has a set of parents $\pi_i$. Let $\phi_i[j]$ denote the $j$th unique instantiation of $\pi_i$ relative to $D$. Suppose there are $q_i$ such unique instantiations of $\pi_i$. Define $\alpha_{ijk}$ to be the number of cases in $D$ in which variable $x_i$ is instantiated as $v_{ik}$ and $\pi_i$ is instantiated as $\phi_i[j]$. Let $N_{ij} = \sum_{k=1}^{r_i} \alpha_{ijk}$. If Assumptions 1 through 5 hold, then

$$P(B_S, D) = P(B_S) \prod_{i=1}^{n} \prod_{j=1}^{q_i} \frac{(r_i - 1)!}{(N_{ij} + r_i - 1)!} \prod_{k=1}^{r_i} \alpha_{ijk}! \cdot \quad (7)$$

$\square$

Equation 7 allows us to calculate $P(B_S, D)$ using knowledge of $P(B_S)$ combined with simple enumeration over the cases in the database. For example, by applying Equation 7 to the structures in Figure 1 with the data in Table 1, we find that $P(B_{S1}, D) = 8.91 \times 10^{-11}$ and $P(B_{S2}, D) = 8.91 \times 10^{-12}$, if we assume uniform priors on $P(B_S)$. Note that these numbers are small because the probability of seeing *exactly* those data that are in $D$ is small. Given the assumptions in this section, the data imply that $B_{S1}$ is 10 times more likely than $B_{S2}$. This result is not surprising, because we used $B_1$ to generate $D$ by the application of Monte Carlo sampling.

Consider the time complexity of computing Equation 7. Let $r = \max_i[r_i]$ for $i = 1$ to $n$. Define $t_{B_S}$ to be the time required to compute the prior probability of structure $B_S$, $P(B_S)$. In [Cooper and Herskovits 1991] we show that the time complexity of computing Equation 7 is $O(m \, n^2 \, r + t_{B_S})$.

Using Equation 7, we can calculate posterior probabilities of belief-network structures as

$$P(B_S \mid D) = \frac{P(B_S, D)}{\sum_{B_S} P(B_S, D)}. \quad (8)$$

Applying Equation 8 to our previous example, we obtain $P(B_{S1} \mid D) = 0.109$, and $P(B_{S2} \mid D) = 0.011$. The remaining probability mass of 0.88 is distributed among the other 23 possible three-node belief-network structures. When there are more than a few variables in the model, the complexity of computing Equation 8 is intractable, due to the large number of belief-network structures. Consider, however, the situation in which

$\sum_{B_S \in Y} P(B_S, D) \approx P(D)$, for some set $Y$ of structures, where $|Y|$ is small. If $Y$ can be efficiently located, then

Equation 8 can be efficiently computed to a close approximation.

## 2.3 FINDING THE MOST PROBABLE BELIEF-NETWORK STRUCTURE

Consider the problem of determining the belief-network structure $B_S$ that maximizes $P(B_S \mid D)$. Knowing such a structure may lend insight into the causal relationships among the model variables, particularly if the structure has a high posterior probability. The structure also may be augmented with numerical probabilities, as we discuss in [Cooper and Herskovits 1991], and used to perform probabilistic inference.

For a given database $D$, $P(B_S, D) \propto P(B_S \mid D)$, and therefore finding the $B_S$ that maximizes $P(B_S \mid D)$ is equivalent to finding the $B_S$ that maximizes $P(B_S, D)$. We can maximize $P(B_S, D)$ by applying exhaustively Equation 7 for every possible $B_S$ containing just the variables in $\mathbf{Z}$. As a function of the number of variables, the number of possible structures grows super-exponentially. Thus, an exhaustive enumeration of all network structures is not feasible in most domains. In particular, Robinson [Robinson 1976] has derived an efficiently computable recursive function for determining the number of possible belief-network structures that contain $n$ nodes. For $n = 2$, the number of possible structures is 3; for $n = 3$, it is 25; for $n = 5$, it is 29,000; and, for $n = 10$, it is approximately $4.2 \times 10^{18}$. Clearly, we need a method that is more efficient than is exhaustive enumeration for locating the $B_S$ that maximizes $P(B_S \mid D)$. In Section 2.3.1, we introduce additional assumptions that reduce the time complexity of enumeration. The complexity, however, remains exponential. Thus, in Section 2.3.2 we introduce and discuss a heuristic method that is polynomial time.

### 2.3.1 An Exhaustive Search Procedure

Let us assume that we can specify an ordering on all $n$ variables, such that if $x_i$ precedes $x_j$ in the ordering, then we do not allow structures in which there is an arc from $x_j$ to $x_i$. In some domains, the time precedence of event variables could be used to establish such an ordering. Given an ordering as a constraint, there remain $2^{\binom{n}{2}} = 2^{n(n-1)/2}$ possible belief-network structures. Let $t(n) = 2^{n(n-1)/2}$. For large $n$, it is not feasible to apply Equation 7 for each of $t(n)$ possible structures. Therefore, in addition to a node ordering, let us assume equal priors on $B_S$. That is, initially, before we observe the data $D$, we believe that all structures are equally likely. In that case, we obtain

$$P(B_S, D) = c \prod_{i=1}^{n} \prod_{j=1}^{q_i} \frac{(r_i - 1)!}{(N_{ij} + r_i - 1)!} \prod_{k=1}^{r_i} \alpha_{ijk}! , \qquad (9)$$

where $c = 1/t(n)$ is our prior probability, $P(B_S)$, for each $B_S$. To maximize Equation 9, it is sufficient to find the parent set of each variable that maximizes the second inner product. Thus, we have

$$\underset{B_S}{max}[\, P(B_S, D)] =$$

$$c \prod_{i=1}^{n} \underset{\pi_i}{max}[ \prod_{j=1}^{q_i} \frac{(r_i - 1)!}{(N_{ij} + r_i - 1)!} \prod_{k=1}^{r_i} \alpha_{ijk}! \,] , \qquad (10)$$

where the maximization takes place over every possible set of parents $\pi_i$ of $x_i$ that is consistent with the ordering on the nodes. A generalization of Equation 10, which is discussed in [Cooper and Herskovits 1991], does not assume that $P(B_S)$ is uniform. Although solving Equation 10 is no longer super-exponential in $n$, it remains exponential in $n$. Thus, further computational improvements are needed.

### 2.3.2 A Heuristic Search Procedure

We propose here one polynomial-time heuristic method, among many possibilities, that attempts to find the $B_S$ that maximizes (or nearly maximizes) $P(B_S \mid D)$. We shall use Equation 10 as our starting point, with the attendant assumptions that we have an ordering on the domain variables and that, a priori, all structures are considered equally likely. We shall modify the maximization operation on the right of Equation 10 to use a greedy-search method. In particular, we use an algorithm that begins by assuming that a node has no parents, and that then adds incrementally that parent whose addition most increases the probability of the resulting structure. When the addition of no single parent can increase the probability, we stop adding parents to the node. We shall use the following function:

$$g(i, \pi_i) = \prod_{j=1}^{q_i} \frac{(r_i - 1)!}{(N_{ij} + r_i - 1)!} \prod_{k=1}^{r_i} \alpha_{ijk}! , \qquad (11)$$

where the $\alpha_{ijk}$ are computed relative to $\pi_i$ being the parents of $x_i$ and relative to a database $D$ which we leave implicit. Let $u$ be the maximum number of parents allowed for any node. In [Cooper and Herskovits 1991] we show that $g(i, \pi_i)$ can be computed in $O(m\, u\, r)$ time. We also shall use a function $\mathrm{Pred}(x_i)$ that returns the set of nodes that precede $x_i$ in the node ordering. Figure 2 contains the heuristic search algorithm, which we call K2. The algorithm is named K2 because it evolved from a system named Kutató [Herskovits and Cooper 1990] that applies the same search heuristics to construct belief networks; Kutató uses entropy to score network structures .

As shown in [Cooper and Herskovits 1991], the time complexity of K2 is $O(m\,u^2\,n^2\,r)$. This result assumes that the factorials we need in order to apply Equation 11 have been precomputed and stored in an array. We can further improve runtime speed by replacing $g(i, \pi_i)$ and $g(i, \pi_i \cup \{z\})$ in K2 by $log(g(i, \pi_i))$ and $log(g(i, \pi_i \cup \{z\}))$, respectively. The logarithmic version of Equation 11 requires only addition and subtraction rather than multiplication and division. If the logarithmic version of Equation 11 is used in K2 then the logarithms of factorials should be precomputed and stored in an array.

1.  **procedure** K2;
2.  {Input: A set of $n$ nodes, an ordering on the nodes, an upper bound $u$ on the number of parents a node may have, and a database $D$ containing $m$ cases.}
    {Output: For each node, a printout of the parents of the node.}
3.  **for** $i := 1$ to $n$ **do**
4.         $\pi_i := \varnothing$;
5.         $P_{old} := g(i, \pi_i)$; {This function is computed using Equation 11.}
6.         OKToProceed := **true**;
7.         **while** OKToProceed and $|\pi_i| < u$ **do**
8.                  let $z$ be the node in Pred($x_i$) - $\pi_i$ that maximizes $g(i, \pi_i \cup \{z\})$;
9.                  $P_{new} := g(i, \pi_i \cup \{z\})$;
10.                 **if** $P_{new} > P_{old}$ **then**
11.                        $P_{old} := P_{new}$;
12.                        $\pi_i := \pi_i \cup \{z\}$
13.                 **else** OKToProceed := false;
14.         **end** {while};
15.         write('Node: ', $x_i$, ' Parents of this node: ', $\pi_i$);
16.  **end** {for};
17.  **end** {K2};

Figure 2: The K2 algorithm heuristically searches for the most probable belief-network structure, given a database of cases and a set of assumptions (see text).

We emphasize that K2 is just one of many possible methods for searching the space of belief networks to maximize the probability metric developed in Section 2.2. Accordingly, the metric developed in Section 2.2 is a more fundamental result than is the K2 algorithm. Nonetheless, K2 has proved valuable as an initial search method for obtaining some preliminary test results, which we shall describe in Section 3.

## 3  PRELIMINARY RESULTS

In this section we describe an experiment in which we generated a database from a belief network by simulation, and then attempted to reconstruct the belief network from the database. In particular, we applied the K2 algorithm to a database of 10,000 cases generated from the ALARM belief network. Beinlich constructed the ALARM network as a research prototype to model potential anesthesia problems in the operating room [Beinlich, Suermondt, et al. 1989]. ALARM contains 46 arcs and 37 nodes, and each node has from two to five possible values. We generated cases using a Monte Carlo technique [Henrion 1988]. Each case corresponds to a value assignment for each of the 37 variables. The Monte Carlo technique is an unbiased generator of cases, in the sense that the probability that a particular case is generated is equal to the probability of the case according to the belief network. We generated 10,000 such cases to create a database that we used as input to the K2 algorithm. We supplied K2 with an ordering on the 37 nodes that is consistent with the partial order of the nodes as specified by ALARM.

From the 10,000 cases, the K2 algorithm constructed a network identical to ALARM, except that one arc was missing and one arc was added. A subsequent analysis revealed that the missing arc is not strongly supported by the 10,000 cases. The extra arc was added due to the greedy nature of the search algorithm. The total search time for the reconstruction was approximately 16 minutes and 38 seconds on a Macintosh II running LightSpeed Pascal version 2.0. We analyzed the performance of K2 when given the first 100, 200, 500, 1000, 2000 and 3000 cases from the same 10,000-case database. Using only 3,000 cases, K2 produced in about 5 minutes the same belief network as when it used the full 10,000 cases.

Although preliminary, these results are encouraging because they demonstrate that K2 can reconstruct a moderate size belief network rapidly from a set of cases using readily available computer hardware. We currently are investigating the extent to which the performance of K2 is sensitive to the ordering of the nodes. We also are exploring methods that do not require an ordering.

## 4  SUMMARY OF THE LEARNING METHOD AND RELATED WORK

In the preceding sections, we have described a Bayesian approach to learning the dependency relationships among a set of discrete variables. For notational simplicity, we shall call the approach BLN (Bayesian learning of belief networks). BLN can represent arbitrary belief-network structures and arbitrary probability distributions on discrete variables. BLN calculates the probability of a structure of variable relationships given a database. The probability of multiple structures can be computed and displayed to the user. BLN also can use multiple structures in performing inference, as we discuss in [Cooper and Herskovits 1991]. When the number of domain variables is large, the combinatorics of enumerating all possible belief network structures becomes prohibitively expensive. Developing better methods for efficiently locating highly probable structures remains an open area of research. BLN is able to represent the prior probabilities of belief-network structures. For example, an expert could attach a high probability to the presence of an arc from node $x$ to node $y$, indicating that—according to current scientific belief—it is very likely that $x$ directly influences $y$. More generally, a prior probability could be specified for the presence of a *set* of arcs. If prior probability distributions on such structures are not available to the computer, then uniform priors can be assumed.

Previously described methods for learning belief networks from databases are non-Bayesian [Chow and Liu 1968, Fung and Crawford 1990, Geiger, Paz, et al. 1990, Herskovits and Cooper 1990, Pearl and Verma 1991, Rebane and Pearl 1987, Spirtes and Glymour 1990, Spirtes, Glymour, et al. 1990, Srinivas, Russell, et al. 1990, Verma and Pearl 1990, Wermuth and Lauritzen 1983]. With non-Bayesian methods, there is no principled way to attach prior probabilities to individual arcs or sets of arcs. In addition, all of these methods, except [Herskovits and Cooper 1990], rely on having threshold values (e.g., p values) for determining when conditional independence holds among variables. BLN does not require the use of such thresholds. Also, non-Bayesian belief-network methods, as well as classical statistical methods, emphasize finding the single most likely structure, which they then may use for inference. They do not, however, quantify the likelihood of that structure. If a single structure is used for inference, implicitly the probability of that structure is assumed to be 1.

### Acknowledgements

We thank Lyn Dupré and Clark Glymour for helpful comments on an earlier draft. This work was supported in part by the National Science Foundation under grant IRI-8703710 and by the U.S. Army Research Office under grant P-25514-EL. Computing resources were provided in part by the SUMEX-AIM resource under grant LM-05208 from the National Library of Medicine.

### References

Agogino, A.M. and Rege, A., IDES: Influence diagram based expert system, *Mathematical Modelling* 8 (1987) 227-233.

Andreassen, S., Woldbye, M., Falck, B. and Andersen, S.K., MUNIN — A causal probabilistic network for interpretation of electromyographic findings, In: *Proceedings of the International Joint Conference on Artificial Intelligence,* Milan, Italy (1987) 366-372.

Beinlich, I.A., Suermondt, H.J., Chavez, R.M. and Cooper, G.F., The ALARM monitoring system: A case study with two probabilistic inference techniques for belief networks, In: *Proceedings of the Conference on Artificial Intelligence in Medical Care,* London (1989) 247-256.

Chow, C.K. and Liu, C.N., Approximating discrete probability distributions with dependence trees, *IEEE Transactions on Information Theory* 14 (1968) 462-467.

Cooper, G.F., Current research directions in the development of expert systems based on belief networks, *Applied Stochastic Models and Data Analysis* 5 (1989) 39-52.

Cooper, G.F. and Herskovits, E.H., A Bayesian method for the induction of probabilistic networks from data, Report SMI-91-1, Section of Medical Informatics, University of Pittsburgh, 1991.

deGroot, M.H., *Optimal Statistical Decisions* (McGraw-Hill, New York, 1970).

Fung, R.M. and Crawford, S.L., Constructor: A system for the induction of probabilistic models, In: *Proceedings of AAAI,* Boston, Massachusetts (1990) 762-769.

Geiger, D., Paz, A. and Pearl, J., Learning causal trees from dependence information, In: *Proceedings of AAAI,* Boston, Massachusetts (1990) 770-776.

Heckerman, D.E., *Probabilistic Similarity Networks,* Ph.D. dissertation, Medical Information Sciences, Stanford University (1990).

Heckerman, D.E., Horvitz, E.J. and Nathwani, B.N., Update on the Pathfinder project, In: *Proceedings of the Symposium on Computer Applications in Medical Care* (1989) 203-207.

Henrion, M., Propagating uncertainty in Bayesian networks by logic sampling. In: Lemmer J.F. and Kanal L.N. (Eds.), *Uncertainty in Artificial Intelligence 2* (North-Holland, Amsterdam, 1988) 149-163.

Henrion, M., An introduction to algorithms for inference in belief nets. In: Henrion M., Shachter R.D., Kanal L.N. and Lemmer J.F. (Eds.), *Uncertainty in Artificial Intelligence 5* (North-Holland, Amsterdam, 1990) 129-138.

Henrion, M. and Cooley, D.R., An experimental comparison of knowledge engineering for expert systems and for decision analysis, In: *Proceedings of AAAI*, Seattle (1987) 471-476.

Herskovits, E.H. and Cooper, G.F., Kutató: An entropy-driven system for the construction of probabilistic expert systems from databases, In: *Proceedings of the Conference on Uncertainty in Artificial Intelligence,* Cambridge, Massachusetts (1990) 54-62.

Holtzman, S., *Intelligent Decision Systems* (Addison-Wesley, Reading, MA, 1989).

Horvitz, E.J., Breese, J.S. and Henrion, M., Decision theory in expert systems and artificial intelligence, *International Journal of Approximate Reasoning* **2** (1988) 247-302.

Kiiveri, H., Speed, T.P. and Carlin, J.B., Recursive causal models, *Journal of the Australian Mathematical Society* **36** (1984) 30–52.

Lauritzen, S.L. and Spiegelhalter, D.J., Local computations with probabilities on graphical structures and their application to expert systems, *Journal of the Royal Statistical Society (Series B)* **50** (1988) 157-224.

Neapolitan, R., *Probabilistic Reasoning in Expert Systems* (John Wiley & Sons, New York, 1990).

Pearl, J., Fusion, propagation and structuring in belief networks, *Artificial Intelligence* **29** (1986) 241-288.

Pearl, J., *Probabilistic Reasoning in Intelligent Systems* (Morgan Kaufmann, San Mateo, California, 1988).

Pearl, J. and Verma, T.S., A theory of inferred causality, In: *Proceedings of the Second International Conference on the Principles of Knowledge Representation and Reasoning,* Boston, MA (1991) 441-452.

Rebane, G. and Pearl, J., The recovery of causal poly-trees from statistical data, In: *Proceedings of the Workshop on Uncertainty in Artificial Intelligence,* Seattle, Washington (1987) 222-228.

Robinson, R.W., Counting unlabeled acyclic digraphs (Note: This paper also discusses counting labeled acyclic graphs.), In: *Proceedings of the Fifth Australian Conference on Combinatorial Mathematics,* Melbourne, Australia (1976) 28-43.

Shachter, R.D., Intelligent probabilistic inference. In: Kanal L.N. and Lemmer J.F. (Eds.), *Uncertainty in Artificial Intelligence* (North-Holland, Amsterdam, 1986) 371-382.

Shachter, R.D., Probabilistic inference and influence diagrams, *Operations Research* **36** (1988) 589-604.

Shachter, R.D. and Kenley, C.R., Gaussian influence diagrams, *Management Science* **35** (1989) 527-550.

Spirtes, P. and Glymour, C., An algorithm for fast recovery of sparse causal graphs, Report CMU-LCL-90-4, Department of Philosophy, Carnegie–Mellon University, 1990.

Spirtes, P., Glymour, C. and Scheines, R., Causal hypotheses, statistical inference, and automated model specification, Unpublished report, Department of Philosophy, Carnegie–Mellon University, 1990.

Srinivas, S., Russell, S. and Agogino, A., Automated construction of sparse Bayesian networks for unstructured probabilistic models and domain information. In: Henrion M., Shachter R.D., Kanal L.N. and Lemmer J.F. (Eds.), *Uncertainty in Artificial Intelligence 5* (North-Holland, Amsterdam, 1990) 295-308.

Suermondt, H.J. and Amylon, M.D., Probabilistic prediction of the outcome of bone-marrow transplantation, In: *Proceedings of the Symposium on Computer Applications in Medical Care* (1989) 208-212.

Verma, T.S. and Pearl, J., Equivalence and synthesis of causal models, In: *Proceedings of the Conference on Uncertainty in Artificial Intelligence,* Cambridge, Massachusetts (1990) 220-227.

Wermuth, N. and Lauritzen, S., Graphical and recursive models for contingency tables, *Biometrika* **72** (1983) 537-552.